\documentclass[14pt, a4paper]{article}
\usepackage[a4paper, left=3cm, right=3cm, top=3cm, bottom=2cm]{geometry}

\usepackage{amssymb}
\usepackage{amsmath}
\usepackage{txfonts}
\usepackage{mathdots}
 \usepackage{graphicx}
\usepackage{caption}
\usepackage{subcaption}
\usepackage{lineno,hyperref}
\usepackage{changepage}
\usepackage{setspace}
\usepackage{algorithmic}
\usepackage[utf8]{inputenc}
\usepackage{enumerate}
\usepackage{cite}
\usepackage{tabularx,booktabs,caption,paralist}
\usepackage[table]{xcolor}
\usepackage{multirow,paralist}
\usepackage[linesnumbered,ruled,vlined]{algorithm2e}
\usepackage{pdflscape}
\usepackage{longtable}
\DeclareGraphicsExtensions{.pdf,.png,.jpg}

\begin{document}

\title{A Novel Plagiarism Detection Approach Combining BERT-based Word Embedding, Attention-based LSTMs and an Improved Differential Evolution Algorithm}

\author{Seyed Vahid Moravvej$^1$, Seyed Jalaleddin Mousavirad$^2$, Diego Oliva $^3$, \\and Fardin Mohammadi$^4$  \\
\ \\
$^1$Department of Computer Engineering, Isfahan University of Technology, Isfahan, Iran \\
$^2$Universidade da Beira Interior, Covilhã, Portugal \\
$^3$Depto. de Innovación Basada en la Información y el Conocimiento, Universidad de Guadalajara,\\ CUCEI, Guadalajara, Mexico\\
$^4$Electrical and Computer Engineering
Department, Semnan University, Semnan, Iran}

\date{}

\maketitle

\begin {abstract}
Detecting plagiarism involves finding similar items in two different sources. In this article, we propose a novel method for detecting plagiarism that is based on attention mechanism-based long short-term memory (LSTM) and bidirectional encoder representations from transformers (BERT) word embedding, enhanced with optimized differential evolution (DE) method for pre-training and a focal loss function for training. BERT could be included in a downstream task and fine-tuned as a task-specific BERT can be included in a downstream task and fine-tuned as a task-specific structure, while the trained BERT model is capable of detecting various linguistic characteristics. Unbalanced classification is one of the primary issues with plagiarism detection. We suggest a focal loss-based training technique that carefully learns minority class instances to solve this. Another issue that we tackle is the training phase itself, which typically employs gradient-based methods like back-propagation for the learning process and thus suffers from some drawbacks, including sensitivity to initialization. To initiate the BP process, we suggest a novel DE algorithm that makes use of a clustering-based mutation operator. Here, a winning cluster is identified for the current DE population, and a fresh updating method is used to produce potential answers. We evaluate our proposed approach on three benchmark datasets ( MSRP, SNLI, and SemEval2014) and demonstrate that it performs well when compared to both conventional and population-based methods.

\textbf{Keywords:} plagiarism detection; LSTM; attention mechanism; BERT; differential evolution; focal loss.
\end {abstract}

\section{Introduction}
\label{sec:Intro}
With an abundance of information easily available online coupled with powerful search engines, plagiarism has become a sensitive issue in various domains including education~\cite{moravvej2021method}. Plagiarism usually occurs intentionally or unknowingly~\cite{sabeeh2021plagiarism}, while plagiarism methods can also be used in other applications such as information retrieval \mbox{~\cite{zhu2020deep,gong2021efficient,zhang2022cbgru,wang2021blockchain,01,02}}, where some text is given as input and the most relevant matches returned.

In the literature, numerous methods for detecting plagiarism have been proposed. Text distance approaches aim to measure the semantic proximity of two textual words in terms of distance. Generally, there are three types of such distances, length distance, distribution distance, and semantic distance~\cite{wang2020measurement}. Length distance methods measure the semantic similarity of two texts based on the numerical characteristics of the text. The most popular methods here are Euclidean distance, cosine distance, and Manhattan distance~\cite{mahmoud2017semantic}. Distance-based methods suffer from two significant problems. They are suitable only for symmetrical problems and limited to problems such as question answering while using distances regardless of statistical characteristics is risky~\cite{deza2009encyclopedia}. Distribution distances, such as Jensen–Shannon divergence~\cite{menendez1997jensen} and Kullback–Leibler divergence~\cite{van2014renyi}, estimate the semantic similarity between two items by comparing their distributions and operate based on lexical and semantic similarities between two texts.  A literature review based on machine learning and deep learning approaches is provided in Table~\ref{Literature}.  Typically, methods based on machine learning suffer from two critical weaknesses. First, they frequently ignore the semantic information contained in keywords, features, or rules, which prevents them from considering the relationships between sentence pairs on all sides. Second, feature extraction and handmade rules lack flexibility, reducing their generalisation ability. Deep learning approaches have superseded earlier methods due to their inherent advantages, including the ability for automated feature extraction{~\cite{moravvej2021biomedical}}. However, the architecture, training algorithms, and embedding choice are primarily responsible for the success of deep models.


Neural network methods, including deep networks, are usually based on gradient-based methods such as backpropagation (BP) to identify the appropriate network weights~\cite{sartakhti2021persian,moravvej2022rlmd,moravvej2021efficient}. Unfortunately, these methods are sensitive to parameter initialization and have the propensity to become trapped in local optima. Initial weights can have a greater effect on neural network performance than network architecture and training examples~\cite{de2016overview}. To overcome these problems, meta-heuristic methods~\cite{vakilian2021using,vakilian2021using1} for example differential evolution (DE)~\cite{storn1997differential} can be used and have been shown to be suitable for optimising the learning process~\cite{moravvej2021lstm,mousavirad2021rde}.

DE is a powerful meta-heuristic algorithm that has been successfully adapted to a variety of optimization problems~\cite{deng2021improved,mousavirad2020evolving}. It comprises three main steps, mutation to create a new candidate solution depending on scaling differences between solutions, crossover to combine the generated mutation vector with the original vector, and selection to choose the best candidate solutions for the next iteration. The operator of the mutation is particularly important~\cite{bajer2019adaptive}.

According to the literature, language model pre-training is considered to have an impact on the improvement of several natural language processing tasks{~\cite{howard2018universal}}. These tasks can be divided into two levels, namely sentence-level and token-level. As their names suggest, the former group of tasks utilises a holistic analysis of sentences to predict the relationships between them. According to{~\cite{williams2017broad}}, natural language inference is an instance of sentence-level tasks, as well as paraphrasing{~\cite{dolan2005automatically}}. However, the latter group, that is, token-level tasks (e. g. named entity recognition and question answering), require models to produce fine-grained output at the token level{~\cite{rajpurkar2016squad}}. Feature-based and fine-tuning strategies are the two approaches that can be implemented to apply pre-trained language representations in order to downstream tasks. On the one hand, feature-based approaches, for instance, ELMo{~\cite{matthew2018peters}}, utilise task-specific architectures that incorporate pre-trained representations as supplementary features. On the other hand, fine-tuning approaches introduce minimal task-specific parameters. These strategies, for example, the Generative Pre-trained Transformer (OpenAI GPT){~\cite{radford2018improving}}, fine-tune all pre-trained parameters to be trained on the downstream tasks. Interestingly, both of these approaches have the same objective function during the process of pre-training. These strategies both aim at learning general language representations by benefiting from unidirectional language models. Current techniques, particularly fine-tuning approaches, limit the efficacy of pre-trained representations, according to the authors of this study. The most significant limitation is that these standard language models are one-way, which limits the architectures that can be used during the pre-training process. For instance, OpenAI GPT requires researchers to use architecture from left to right; hence, in the self-attention layers of the Transformer, every token is bound to attend to the previous tokens only{~\cite{vaswani2017attention}}. Limitations such as the one mentioned above are sub-optimal for sentence-level tasks. Furthermore, if approaches based on fine-tuning are applied to token-level tasks (e. g., question answering), due to the significant importance of incorporating context from both directions in such tasks, these limitations can be very harmful. This current study aims at fostering approaches based on fine-tuning by proposing BERT: Bidirectional Encoder Representations from Transformers. Inspired by the Cloze task{~\cite{taylor1953cloze}}, BERT can alleviate the previously-mentioned constraints by utilising a masked language model (MLM) pre-training objective. By masking some of the tokens from the input randomly, the masked language model objective is to use the masked word contexts to predict their original vocabulary id. Contrary to the left-to-right language model pre-training, the left and the right contexts can be fused by the MLM objective, allowing a deep bidirectional Transformer to be pre-trained. Furthermore, we use a next-sentence prediction task, which jointly pre-trains text-pair representations, as well.

In this paper, we introduce a novel attention mechanism-based LSTM model for plagiarism detection founded on BERT word embedding and a clustering-based DE algorithm. Our proposed BPD-IDE model includes two LSTMs for source and suspicious sentences as well as a feed-forward network for predicting their similarity. Positive and negative pairs are used to train the model, with positive pairs consisting of two similar sentences and negative pairs of two dissimilar sentences. We use BERT word embedding to discover the semantic similarity between sentences without the use of pre-engineered features. Importantly, we present an enhanced DE algorithm based on clustering for weight initialization in order to discover a promising region in search space to start the BP algorithm in all LSTMs, the attention mechanism, and the feed-forward network. For this, the best candidate solution in the best cluster is selected as the initial solution in the mutation operator, and a novel updating strategy is employed to create candidate solutions. In addition, our proposed algorithm employs focal loss (FL)~\cite{Focal_Loss} to class imbalance. We assess our BPD-IDE model using three reference datasets, MSRP, SNLI and SemEval2014, and show BPD-IDE to be superior to other approaches. The present study has three main contributions:

\begin{itemize}
	\item We introduce BERT word embedding, the most recently developed model for several languages, for plagiarism detection.
	\item We propose a novel DE algorithm to intialise weight values
	\item We propose a focal loss-based training approach for plagiarism detection, which results in the prevention of problems regarding the imbalance classification by learning minority class instances.
	
\end{itemize}

The remainder of the paper is structured as following. Section~\ref{sec:background} covers some background knowledge, while Section~\ref{sec:proposed} introduces our proposed plagiarism detection method. Section~\ref{Sec:exp} presents experimental results, and Section~\ref{Sec:conc} concludes the paper.

\begin{landscape}
	\label{Literature}
	\begin{longtable}{lcccccc}
		
		Ref. & Technique/ Algorithm  & Dataset source & Method & Result \\
		\hline			
		~\cite{menai2012detection} & \begin{tabular}{@{}c@{}}New comparison algorithm\\ that uses heuristics \end{tabular} & \begin{tabular}{@{}c@{}}300 Arabic documents \\ available on Alwaraq website \end{tabular} & \begin{tabular}{@{}c@{}}Fingerprinting and Similarity \\ metric with Arabic WordNet \end{tabular} & Accuracy: 93\% 
		\\ \hline

		~\cite{jadalla2012plagiarism} & 
		\begin{tabular}{@{}c@{}}Winnowing ngram\\ fingerprinting algorithm \end{tabular} & 
		\begin{tabular}{@{}c@{}}3 dataset\\ a set of students' projects (131 documents)\\ 116,011 Arabic Wikipedia files (386 MB) \\ manually compiled files \end{tabular}
		& 
		\begin{tabular}{@{}c@{}}Relative Frequency Model (RFM) \end{tabular}
		& 
		Accuracy: 94\% 
		\\ \hline

		~\cite{hussein2015plagiarism} & 
		\begin{tabular}{@{}c@{}}Heuristic pairwise phrase matching \\ algorithm and TF-IDF model \\ and 300 Arabic documents\end{tabular} & 
		\begin{tabular}{@{}c@{}}3,300 Arabic documents \end{tabular}
		& 
		\begin{tabular}{@{}c@{}}Latent Semantic Analysis and \\ the Singular Value Decomposition (SVD) \end{tabular}
		& 
		Accuracy: 85.9\%
		\\ \hline

		~\cite{pennington2014glove} & 
		\begin{tabular}{@{}c@{}}The GloVe Model \end{tabular} & 
		\begin{tabular}{@{}c@{}}CoNLL-2003 Shared Task \end{tabular}
		& 
		\begin{tabular}{@{}c@{}} recursive neural networks (RNN) \end{tabular}
		& 
		Accuracy: 75\% 
		\\ \hline

		~\cite{eyecioglu2015twitter} & 
		\begin{tabular}{@{}c@{}}Simple Overlap Features and SVMs \end{tabular} & 
		\begin{tabular}{@{}c@{}}Twitter paraphrases \end{tabular}
		& 
		\begin{tabular}{@{}c@{}}SVM \end{tabular}
		& 
		Accuracy: 86.5\% 
		\\ \hline

		~\cite{khemakhem2016iso} & 
		\begin{tabular}{@{}c@{}}Fuzzy-set IR model \end{tabular} & 
		\begin{tabular}{@{}c@{}}100 documents with 4,367 statement \end{tabular}
		& 
		\begin{tabular}{@{}c@{}} Term-to-term correlation factor and  \\ term-tosentence correlation \end{tabular}
		& 
		Accuracy: 78\% 
		\\ \hline

		~\cite{franco2016cross} & 
		\begin{tabular}{@{}c@{}}Multilingual semantic network \end{tabular} & 
		\begin{tabular}{@{}c@{}}PAN-PC-2011 \end{tabular}
		& 
		\begin{tabular}{@{}c@{}} Artificial neural networks \end{tabular}
		& 
		Accuracy: 75\% 
		\\ \hline

		~\cite{suleiman2017deep} & 
		\begin{tabular}{@{}c@{}}Word2vec model \end{tabular} & 
		\begin{tabular}{@{}c@{}}OSAC corpus \end{tabular}
		& 
		\begin{tabular}{@{}c@{}} Cosine similarity measure \end{tabular}
		& 
		Accuracy: 99\% 
		\\ \hline

		~\cite{pontes2018predicting} & 
		\begin{tabular}{@{}c@{}}CNN + LSTM \end{tabular} & 
		\begin{tabular}{@{}c@{}}SICK dataset \end{tabular}
		& 
		\begin{tabular}{@{}c@{}} CNN + LSTM \end{tabular}
		& 
		Accuracy: 85.49\% 
		\\ \hline

		~\cite{chen2018rnn} & 
		\begin{tabular}{@{}c@{}} CA-RNN \end{tabular} & 
		\begin{tabular}{@{}c@{}}TREC-QA  \&   WikiQA   \&   MSRP\end{tabular}
		& 
		\begin{tabular}{@{}c@{}} CA-RNN \end{tabular}
		& 
		Accuracy: 77.3\% 
		\\ \hline

		~\cite{bao2018attentive} & 
		\begin{tabular}{@{}c@{}}Attention mechanism \end{tabular} & 
		\begin{tabular}{@{}c@{}}SICK dataset \end{tabular}
		& 
		\begin{tabular}{@{}c@{}} LSTM \end{tabular}
		& 
		Accuracy: 78\% 
		\\ \hline

		~\cite{wali2017using} & 
		\begin{tabular}{@{}c@{}}New algorithm that uses \\semantic sentence similarity \end{tabular} & 
		\begin{tabular}{@{}c@{}}A corpus of 300 student thesis reports \end{tabular}
		& 
		\begin{tabular}{@{}c@{}}Novel method to compute semantic similarity \\between sentences using LMF \\standardized Arabic dictionary ElMadar \end{tabular}
		& 
		Accuracy: 94\% 
		\\ \hline

		~\cite{khorsi2018two} & 
		\begin{tabular}{@{}c@{}} Fingerprinting and Word embedding \end{tabular} & 
		\begin{tabular}{@{}c@{}} External Arabic Plagiarism Corpu\\ (ExAra- 2015)\end{tabular}
		& 
		\begin{tabular}{@{}c@{}} Measuring the similarity between \\ two using the Jaccard and Cosine similarities \end{tabular}
		& 
		Accuracy: 86\% 
		\\ \hline

		~\cite{chi2018sentence} & 
		\begin{tabular}{@{}c@{}} Attention-based Siamese network \end{tabular} & 
		\begin{tabular}{@{}c@{}}Disorder Set.\end{tabular}
		& 
		\begin{tabular}{@{}c@{}} LSTM + FNN + attention \end{tabular}
		& 
		Accuracy: 86.1\% 
		\\ \hline

		~\cite{laskar2020contextualized} & 
		\begin{tabular}{@{}c@{}} RoBERTa model \end{tabular} & 
		\begin{tabular}{@{}c@{}}The YahooCQA dataset\end{tabular}
		& 
		\begin{tabular}{@{}c@{}} ELMo + BERT \end{tabular}
		& 
		Accuracy: 95.5\% 
		\\ \hline

		~\cite{moravvej2021method} & 
		\begin{tabular}{@{}c@{}} LSTM-AM \end{tabular} & 
		\begin{tabular}{@{}c@{}} SemEval2014, STS  and MSRP \end{tabular}
		& 
		\begin{tabular}{@{}c@{}} BLSTM \end{tabular}
		& 
		Recall: 95.27\%, 97.41\% and 97.37\%
		\\ \hline

		~\cite{moravvej2022improved} & 
		\begin{tabular}{@{}c@{}} BERT+ LSTM + DE \end{tabular} & 
		\begin{tabular}{@{}c@{}}SNLI, MSRP and SemEval2014\end{tabular}
		& 
		\begin{tabular}{@{}c@{}}  LSTM + BERT \end{tabular}
		& 
		Accuracy: 91.9\%, 92.4\%, 86.5\% 
		\\ \hline

		~\cite{moravvej2021lstm} & 
		\begin{tabular}{@{}c@{}} Pre-Training Parameters \end{tabular} & 
		\begin{tabular}{@{}c@{}}STS Semantic, MSRP and SemEval2014\end{tabular}
		& 
		\begin{tabular}{@{}c@{}} LSTM-based \end{tabular}
		& 
		Accuracy: 97\% 
		\\ \hline

		\caption{List of machine learning and deep learning related works}	
		
		\centering
		
		\label{r}
		
	\end{longtable}
	
\end{landscape}


\section{Background}
\label{sec:background}

\subsection{Long Short-Term Memory (LSTM)}
A type of neural network that is used for sequence-based applications like speech recognition and video processing is a recurrent neural network (RNN)~\cite{medsker2001recurrent}. The network has a layer of feedback in which output is returned with the subsequent input. In every step $t$, the hidden layers are calculated as
\begin{equation}
h_t=H(W_{hh}h_{t-1}+W_{xh}x_t+b_h) ,
\end{equation}
and the output layer as 
\begin{equation}
y_t=H(W_{hy}h_{t-1}+b_y) ,
\end{equation}
Where $W$ and $b$ represent the weight matrix and bias term, respectively, and $H$ is the recurrent hidden layer function. In an RNN, if the sequence is long, the gradients vanish or explode, causing the model to train very gradually~\cite{manaswi2018rnn}.

A long short-term memory (LSTM) network~\cite{hochreiter1997long} is a special type of RNN that can deal with correlations in both short- and long-term sequences by considering the hidden layer as a memory unit. An LSTM memory cell comprises three gates, the input gate $i_t$, the forget gate $f_t$, and the output gate $o_t$. The state of the cell is $h_t$, the input of every gate is the data $x_t$, and the previous state of the memory cell is $h_{t-1}$.


Th input gate is updated as
\begin{equation}
i_t=\sigma(W_ix_t+U_ih_{t-1}+b_i) ,
\end{equation}
the forget gate as
\begin{equation}
f_t=\sigma(W_fx_t+U_fh_{t-1}+b_f) ,
\end{equation}
and the output gate as
\begin{equation}
o_t=\sigma(W_ox_t+U_oh_{t-1}+b_o) ,
\end{equation}
where $\sigma$ is an activation function. The state update is then calculated as
\begin{equation}
h_t=o_t\tanh(c_t) ,
\end{equation}
with
\begin{equation}
c_t=f_tc_{t-1}+i_t\tanh(W_jx_t+U_jh_{t-1}+b_j) .
\end{equation}

A bi-directional LSTM (BLSTM) extends an LSTM network to process input from both sides. This can be useful in plagiarism detection since the user may generate a new sentence by moving the words in the source sentence. In a BLSTM network, the state vectors $\overrightarrow{h_t}$  and $\overleftarrow{h_t}$ are generated by parsing the input, and combines them as $h_t=[\overrightarrow{h_t},\overleftarrow{h_t}]$.

LSTM and BLSTM  networks assume the same importance to each input, leading to network confusion. Consider an attention mechanism as a solution to this issue. For this purpose, for every state $h_t$, the coefficient of attention $\alpha_t$ is attributed, and the final state $h$ for a sequence of length $T$ is computed as
\begin{equation}
h= \sum_{t=1}^{T} \alpha_th_t .
\end{equation}

\subsection{Differential Evolution}
Differential evolution~\cite{storn1997differential} is a population-based algorithm that has been demonstrated to work well for various optimisation problems~\cite{das2009automatic,fister2020post}. DE starts with an initial population, typically derived from a uniform distribution and containing three primary operations: mutation, crossover, and selection.

The mutation operator builds a mutant vector as
\begin{equation} 
\overrightarrow{v}_{i,g}=\overrightarrow{x}_{r_1,g}+F\ (\overrightarrow{x}_{r_2,g}-\overrightarrow{x}_{r_3,g}) ,
\end{equation}
where $\overrightarrow{x}_{r_1,g}$, $\overrightarrow{x}_{r_2,g}$ and $\overrightarrow{x}_{r_3,g}$ are three (different) candidate solutions randomly chosen from the current population, and $F$ is a factor scaling.

Crossover combines the vectors of the mutant and the target. A popular crossover operator is the binomial crossover, which achieves this as
\begin{equation}
u_{i,j,g} =
\begin {cases}
v_{i,j,g} & \text{if } rand(0,1) \le CR \;\; or \;\; j=j_{rand}\\ 
x_{i,j,g} & \text{otherwise}  
\end {cases} ,
\end{equation}
where $CR$ is the crossover rate, and $j_{rand}$ is a random number selected from $\{1, 2,..., D\}$, with $D$ the dimensionality of a candidate solution.

The selection operator then chooses the better solution from the trial and target vectors.

\section{Proposed Approach}
\label{sec:proposed}
\begin{figure}[!t]
	\centering
	\includegraphics[width=\textwidth]{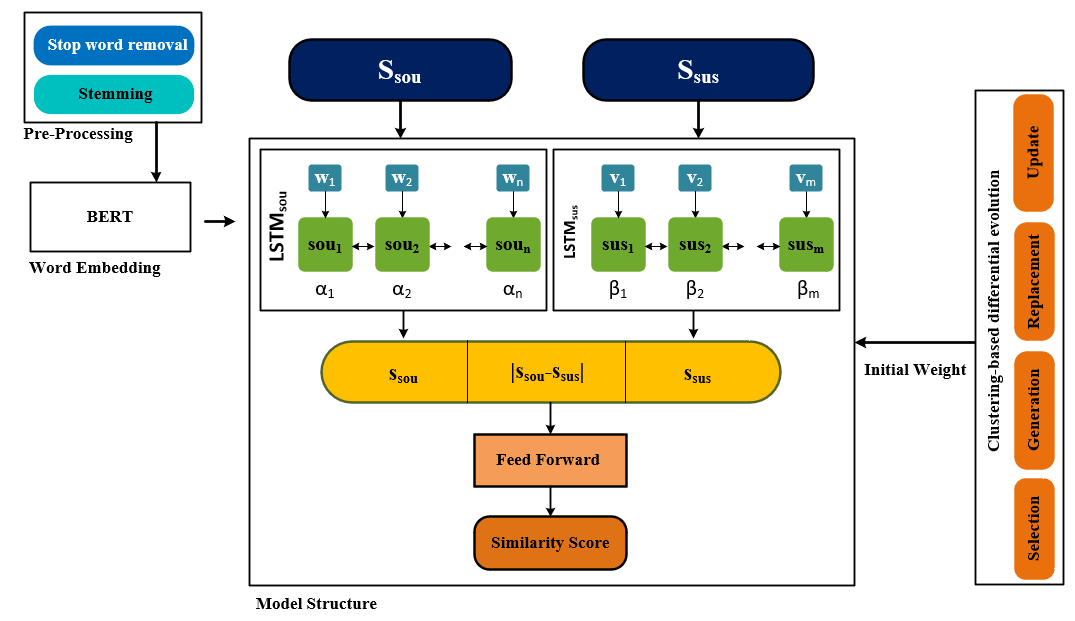}
	\caption{Overall architecture of BPD-IDE model.}
	\label{fig:model}
\end{figure}

Figure~\ref{fig:model} depicts the general architecture of our proposed BPD-IDE approach. As we can see, it comprises three main stages, pre-processing, word embedding, and prediction. First, redundant words and symbols are removed from the sentences. Then, the embedding vector of each word is extracted using BERT, and finally, the similarity of the two sentences is predicted. BPD-IDE incorporates a clustering-based differential evolution algorithm to find the initial seeds of the network weights, while using focal loss to handle class imbalance.



\subsection{Pre-Processing}
Pre-processing is a vital part of any NLP system because the essential characters, words, and sentences identified in this stage are passed to the later stages and thus significantly affect the final result, while an inappropriate pre-processing technique can decrease model performance~\cite{vijayarani2015preprocessing}. We use common stop word elimination and stemming techniques in our approach.


Stop words are part of sentences that can be regarded as overhead. The most common stop words are articles, prepositions, pronouns, etc. They should thus be removed as they cannot function as keywords in text mining applications~\cite{porter1980algorithm} and to reduce the dimensionality of the term space

Stemming is used to identify the stem of a word. For instance, the terms `watch', `watched', `watching', `watcher', etc.\ all can be stemmed to the word ``watch''. Stemming removes ambiguity, reduce the number of words, and reduces time and memory requirements~\cite{vijayarani2015preprocessing}. 	

\subsection{BERT-Based Word Embedding}
The objective of word embedding~\cite{bengio2000neural} is to map words to semantic vectors to be used in algorithms of machine learning. It has been demonstrated to be a reliable method for extracting meaningful word representations based on their context~\cite{liu2018neural}. 
Diverse word embedding techniques, including Skip-gram~\cite{mccormick2016word2vec} and matrix factorization techniques such as GloVe~\cite{pennington2014glove}, have been suggested to produce meaningful word representations for neural network models.


Pre-trained language models (PLMs) typically employ unlabelled data to learn model parameters~\cite{alatawi2021detecting}. In this article, we employ the BERT model~\cite{devlin2018bert} as one of the most recent PLM approaches. BERT is a bi-directional language model trained on large datasets such as Wikipedia to generate contextual representations, and is typically fine-tuned from a neural network dense layer for various classification tasks. The fine-tuning allows to exploit the context or problem-specific meaning with a pre-trained generic meaning and trains it for a classification problem.

\begin{figure}[!t]
	\centering
	\includegraphics[width=0.5\linewidth]{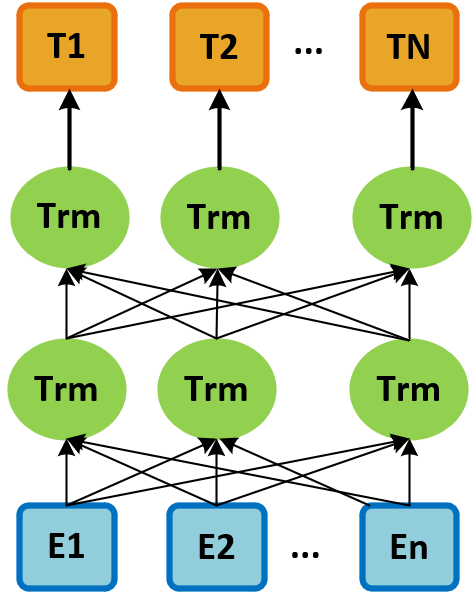}
	\caption{Architecture of the BERT model.}
	\label{bert}
\end{figure}

The general BERT architecture is shown in Figure~\ref{bert}. BERT uses a bi-directional transformer, in which representations are jointly conditioned on both the left and right context in all layers~\cite{devlin2018open}. This distinguishes BERT from Word2Vec and GloVe models which produce an embedding in one direction to ignore its contextual differences.

\section{Prediction}
Our prediction model comprises two attention-based BLSTM networks as extractors of the embeddings of the source and suspicious sentences and a feed-forward network as a predictor of the similarity of the two sentences. Given $S_{sou}=\{w_1,w_1,...,w_n\}$ and $S_{sus}=\{v_1,v_1,...,v_m\}$ as the source and suspicious sentence, where $w_i$ and $v_i$ denote the $i$-th word in the source and suspicious sentence, respectively. $S_ {sou}$ and $S_ {sus}$ are limited to $n$ and $m$ words due to the length limitation in BLSTM (in our work, $n=m$). $S_{sou}$ and $S_{sus}$ are fed separately into an LSTM network. The embeddings of these sentences are computed, using the attention mechanism, as
\begin{equation}
s_{sou}= \sum_{i=1}^{n} \alpha_ih_{sou_i}
\end{equation}
and
\begin{equation}
s_{sus}= \sum_{i=1}^{m} \beta_ih_{sus_i} ,
\end{equation}
where $h_{sou_i}=[\overrightarrow{h}_{sou_i},\overleftarrow{h}_{sou_i}]$ and $h_{sus_i}=[\overrightarrow{h}_{sus_i},\overleftarrow{h}_{sus_i}]$ represent the $i$-th hidden vectors in the BLSTM, and  $\alpha_i, \beta_i \in [0,1]$ are the $i$-th attention weighst for each unit in the BLSTM, calculated as
\begin{equation} 
\alpha_i= \frac{e^{u_i}}{\sum_{j=1}^{n}{e^{u_j}}}
\end{equation}
and
\begin{equation} 
\beta_i= \frac{e^{v_i}}{\sum_{j=1}^{m}{e^{v_j}}} ,
\end{equation}
with
\begin{equation}
u_i= \tanh(W_uh_{sou_i}+b_u)
\end{equation}
and
\begin{equation}
v_i= \tanh(W_vh_{sus_i}+b_v) ,
\end{equation}
where $W_u$ and $W_v$ and $b_u$ and $b_v$ are the weight matrices and biases for the attention mechanisms. The input of the feed-forward network is 



the connection of the $s_{sou}$, $s_{sus}$ and 
$|s_{sou}-s_{sus}|$ as shown in Figure~\ref{fig:model}. The dataset used for training consists of positive and negative pairs, where positive pairs contain a source sentence and a copied sentence, and negative pairs comprise a source sentence and a different sentence.

Our model has two training phases, pre-training and fine-tuning. In pre-training, an appropriate starting configuration is found. The weights obtained in pre-training are then the initial weights of the fine-tune phase. It is in the pre-training phase, where we employ our enhanced differential evolution algorithm.

\subsection{Pre-Training}
At this stage, the weights of the LSTM, attention mechanism, and feed-forward neural network are initialised. For this, we introduce an enhanced differential evolution method that is boosted by a clustering scheme and a novel fitness function.

\subsubsection{Clustering-based differential evolution}
In our enhance DE algorithm, we employ a clustering-based mutation and updating scheme to improve the optimisation performance. The pseudo-code of the algorithm is given in Algorithm~\ref{alg1}.

\begin{algorithm}
	\DontPrintSemicolon	
	\caption{Clustering-based differential evolution algorithm.}
	\label{alg1}
	\textbf{Inputs:} $D$: dimensionality of solution; $MaxFES$: maximum number of function evaluations; $F$: scalling factor; $CR$: crossover probability
	Initialise population $P = (\overrightarrow{x}_1, \overrightarrow{x}_2, \overrightarrow{x}_3 ,..., \overrightarrow{x}_{NP} )$ randomly\\
	Calculate fitness of every solution in $P$\\
	\While{$FES$<$MaxFES$} {
		\For{$i= 1$ to $NP$} {
			Choose three individuals $\overrightarrow{x}_{r_1,g}$, $\overrightarrow{x}_{r_2,g}$, $\overrightarrow{x}_{r_3,g}$ from $P$ randomly, so that $\overrightarrow{x}_{r_1,g} \ne \overrightarrow{x}_{r_2,g} \ne\overrightarrow{x}_{r_3,g}$\\
			$\overrightarrow{v}_{i,g}$=$\overrightarrow{x}_{r_1,g}$+$F(\overrightarrow{x}_{r_2,g}-\overrightarrow{x}_{r_3,g})$\\
			Select $j_{rand}$ as random number in $[0,1]$\\
			\For{$j= 0$ to $D$} {
				\If{$rand(0,1) \le CR$ or $j=j_{rand}$} {
					$u_{i,j,g} = v_{i,j,g}$
				}
				\Else {
					$u_{i,j,g} = x_{i,j,g}$
				}
			}
			\If{$f(\overrightarrow{u}_{i,g})<f(\overrightarrow{x}_{i,g})$} {
				$\overrightarrow{x}_{i,g+1} = \overrightarrow{u}_{i,g}$
			}
			\Else {
				$\overrightarrow{x}_{i,g+1} = \overrightarrow{x}_{i,g}$
			}
		}
		Select $k$ as random number in $[2,\sqrt{N}]$\\
		Cluster $P$ into $k$ clusters\\
		Compute mean fitness of every cluster\\
		Select winner cluster as cluster with lowest mean fitness\\
		$\overrightarrow{win_g}$ = best solution in winner cluster\\
		\For{$j= 1$ to $M$} {
			Choose two solutions $\overrightarrow{x}_{r_1,g}$, $\overrightarrow{x}_{r_2,g}$ from $P$ randomly, so that $\overrightarrow{x}_{r_1,g} \ne \overrightarrow{x}_{r_2,g}$\\
			$\overrightarrow{v^{clu}}_{i,g}= \overrightarrow{win_g} + F(\overrightarrow{x}_{r_1,g}-\overrightarrow{x}_{r_2,g})$\\
		}
		Choose $M$ candidate solutions randomly from $P$ as set $B$\\
		Select best $M$ solutions from $v^{clu}$ as set $B^\prime$\\
		Generate new population as $(P - B) \cup B^\prime$\\
	}
\end{algorithm}

\begin{figure}[h]
	\centering
	\includegraphics[width=.5\linewidth]{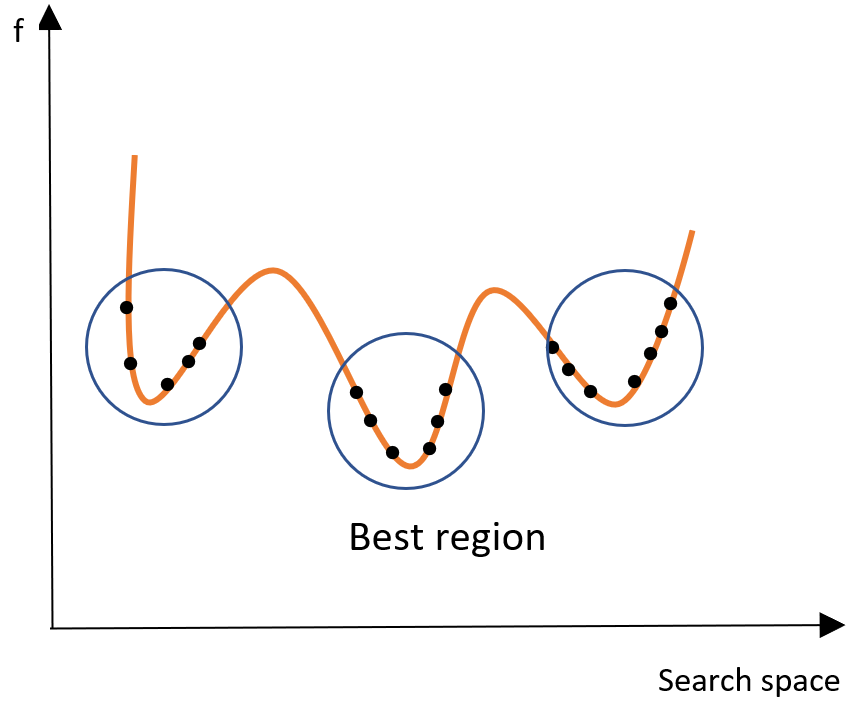}
	\caption{Population clustering in search space to find the best region.}
	\label{points}
\end{figure}

The proposed mutation operator, inspired by~\cite{mousavirad2017human}, identifies a promising region in search space. For this, the current population $P$ is divided into $k$ clusters, using the $k$-means clustering algorithm, so that each cluster represents a region in search space. The number of clusters is selected randomly from $[2,\sqrt{N}]$~\cite{cai2011clustering,mousavirad2021rde}. After clustering, the best cluster is identified as that of the lowest mean fitness of its samples (for a minimisation problem). Figure~\ref{points} shows an example of this process for a toy problem with 18 candidate solutions divided into three clusters.

The proposed clustering-based mutation is defined as
\begin{equation}
\overrightarrow{v^{clu}}_{i}= \overrightarrow{win_{g}} + F(\overrightarrow{x}_{r_1}-\overrightarrow{x}_{r_2}) ,
\end{equation}
where $\overrightarrow{x}_{r_1,g}$ and $\overrightarrow{x}_{r_2,g}$ are two randomly chosen candidate solutions from the current population, and $\overrightarrow{win_g}$ is the best solution in the promising region. Note that $\overrightarrow{win_g} $ is not necessarily the best solution in the population. 

After generating $M$ new solutions by clustering-based mutation, the current population is updated based on the generic population-based algorithm (GPBA)~\cite{deb2005population}, proceedings as follows
\begin{itemize}
	\item
	\textbf {Selection:}  generate $k$ individuals randomly as initial seeds of $k$-means algorithm;
	\item
	\textbf {Generation:} generate $M$ solutions using clustering-based mutation as set $v^{clu}$;
	\item
	\textbf {Replacement:} select $M$ solutions randomly from the current population as set $B$;
	\item
	\textbf {Update:} select the best $M$ solutions from $ v^{clu} \cup B$ as set $B^\prime$. Finally, the new population is obtained as $(P-B) \cup B^\prime$.
\end{itemize}

\subsubsection{Encoding strategy}
\begin{figure}[h]
	\centering
	\includegraphics[width=.9\textwidth]{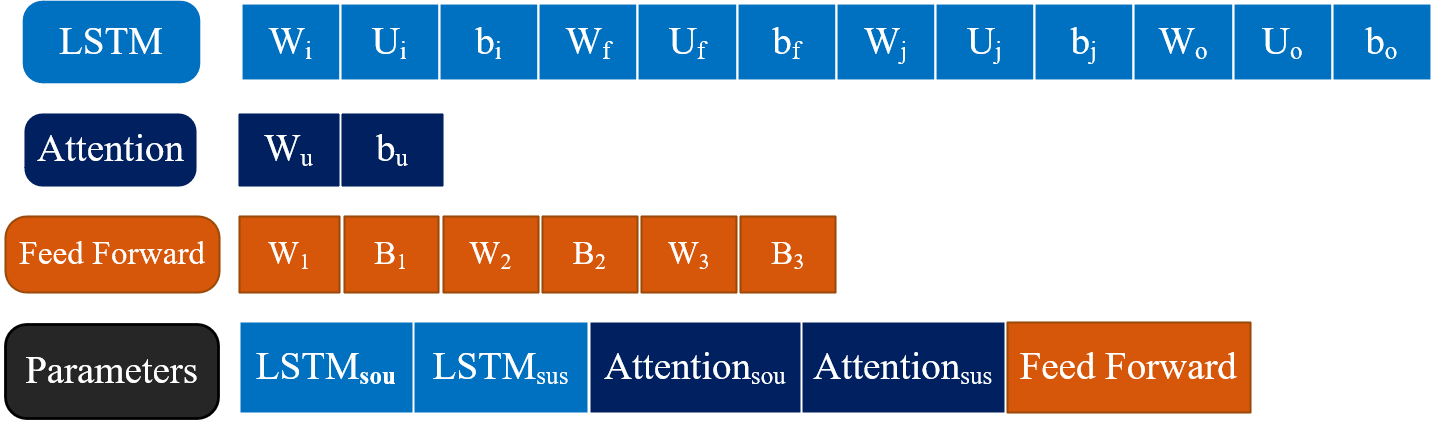}
	\caption{Encoding strategy in our proposed algorithm.}
	\label{weight}
\end{figure}

The primary structure of our proposed model includes two LSTM networks along with their attention mechanisms and a feed-forward network. As illustrated in Figure~\ref{weight}, we arrange all weights and bias terms into a vector to form a candidate solution in our proposed DE algorithm.

\subsubsection{Fitness function}
To calculate the quality of a candidate solution, we define our fitness function as


\begin{equation}
F= \frac{1}{\epsilon + \sum_{i=1}^{N}{(y_i-\tilde{y}_i)^2}} ,
\end{equation}
where $N$ is the total number of training samples, $y_i$ and $\tilde{y}_i$ are the $i$-th target and output predicted by the model, respectively, and  $\epsilon$ is a small number to avoid division by 0. 

\subsection{Focal Loss}
We define the plagiarism problem as a two-class classification problem based on positive and negative classes. As is is an imbalanced problem, with much few samples in the negative class, we use focal loss (FL)~\cite{Focal_Loss} to address this.

FL is a modification of binary cross-entropy (CE) that focusses training on harder (i.e., minority class) samples~\cite{prabhakar2021performance}. CE is defined as 
\begin{equation}
CE =
\begin{cases}
-\log(p) & \text{if } y=1 \\
-\log(1-p) & \text{otherwise}
\end{cases} ,
\end{equation}
where $y \in \{-1,1\}$ is the ground truth class, and $p \in [0, 1]$ is the predicated probability of the model for the class with label $y = 1$. The probability is
\begin{equation}
p_t = 
\begin{cases}
-p & \text{if } y=1 \\
1-p & \text{otherwise}
\end{cases} ,
\end{equation}
and hence we get
\begin{equation}
CE (p,y) = CE(p_t)=-log(p_t) .
\end{equation}

Focal loss adds a modulating factor to cross-entropy loss, leading to
\begin{equation}
FL (p_t) = -\alpha_t(1-p_t)^ \gamma log(p_t)
\end{equation}
where $\gamma > 0$ (if $\gamma = 1$, then FL is similar to CE loss), and $\alpha \in [0,1]$ is the inverse class frequency.

\section{Experimental Results}
\label{Sec:exp}
To demonstrate the effectiveness of our proposed algorithm, we conduct a series of experiments. 

\subsection{Datasets}
In our experiments, we use the following three benchmark datasets:
\begin{itemize}
	\item
	\textbf{SNLI:} the Stanford Natural Language Inference (SNLI) corpus~\cite{bowman2015large} is a large dataset consisting of pairs of labelled sentences with three classes (entailment, contradiction, and semantic independence). It comprises 550,152 sentence pairs for training  and 10,000 sentence pairs each for validation and testing.
	\item
	\textbf{MSRP:} the Microsoft Research Paraphrase Corpus (MSRP)~\cite{chen2018bq} is a paraphrasing dataset of Internet news articles, divided into training and testing, with pairs of sentences labelled as positive and negative by several experts. Of the whole collection, about 67\% of paraphrases are present. The training and test datasets have 4,076 and 1,726 samples, respectively, of which the number of paraphrases is 2,753 and 1,147, respectively.
	\item
	\textbf{SemEval2014:} the Semantic Evaluation Database (SemEval)~\cite{bao2018attentive} is a popular evaluator for STS work, presented in various versions. The sentences from the Com-positional Knowledge (SICK) dataset~\cite {marelli2014sick}, published in 2014, are used to evaluate the semantic relevance of sentences. The dataset consists of 10,000 pairs of sentences broken down into 4,500, 500 and 5,000 pairs for training, validation and testing. Each pair is labelled with a number in the range [1,5], with 1 indicating the lowest degree of similarity, and 5 the highest.
	degree of similarity with number 5. We put those with label 1 in one class, and the rest in another.
\end{itemize}

\subsection{Metrics}
Based on the usual definitions of true positives ($TP$), true negatives ($TN$), false negatives ($FN$), and false positives ($FP$, we can the accuracy
\begin{equation}
Accuracy= \frac{TP+TN}{TP+TN+FP+FN} ,
\end{equation}
recall
\begin{equation}
Recall = \frac{TP}{TP+FN} ,
\end{equation}
and precision
\begin{equation}
Precision = \frac{TP}{TP+FP} .
\end{equation}
In addition, we use the F-measure defined as
\begin{equation}
F-measure = \frac{2 \times Recall\times Precision}{Recall+Precision} ,
\end{equation}
and the G-means
\begin{equation}
G-means = \sqrt{Recall \times Specificity} ,
\end{equation}
with
\begin{equation}
Specificity = \frac{TN}{TN+FP} ,
\end{equation}
which are better suited to evaluate imbalanced classification problems~\cite{lin2020deep}.

\subsection{Model Performance}
Our project employs a 64-bit Windows operating system with 64 GB of RAM and a 64 GB GPU. After 220, 150, and 80 epochs, the best model was obtained for the SNLI, MSRP, and SemEval2014. The total training time for the three datasets was four, three, and one and a half hours.

In the first experiment, our algorithm is compared to seven deep learning-based methods., namely RNN~\cite{sanborn2015deep}, Siamese CNN+LSTM~\cite{pontes2018predicting}, CA-RNN~\cite{chen2018rnn}, AttSiaBiLSTM~\cite{bao2018attentive}, LSTM+FNN+attention~\cite{chi2018sentence}, and CETE~\cite{laskar2020contextualized}.

The results are given in Tables~\ref{r1}, \ref{r2}, and~\ref{r3} for SNLI, MSRP, and SemEval2014, respectively. For our proposed method, we show results based on random weight initialisation, with the use of focal loss, and the full BPD-IDE model. For the SNLI dataset, the BPD-IDE model has beaten other models, including CETE, the best competitor, in all criteria, so that our model has reduced the error by more than 43\% and 44\% in two main criteria, including, F-measure and G-means. 
	By comparing BPD-IDE with \textit{BPD+random weights}, \textit{BPD+random weights+FL}, we can see that it decreases the error rate by about 62\%, indicating the importance of improved DE and FL approaches. For the MSRP
	dataset, BPD-IDE obtained the highest improvement followed by CETE algorithm. The error improvement rate in this database is approximately 25.39\% and 26.47\% for both F-measure and G-means criteria, respectively. In the SemEval2014 dataset, BPD-IDE decreases the classification error by more than 14\% and 31\% compared to CETE and STS-AM, respectively.

\begin{table}[ht]
	\centering
	\caption{Results of deep learning models on SNLI dataset.}
	\begin{tabular}{lccccc}
		\hline
		\hline
		& accuracy & recall & precision & F-measure & G-means \\
		\hline
		\hline
		RNN~\cite{sanborn2015deep}&  0.687&	0.594&	0.540&	0.566&		0.661	\\
		Siamese CNN+LSTM~\cite{pontes2018predicting} &  	0.850 &	0.763&	0.792&	0.777& 0.826	\\
		CA-RNN~\cite{chen2018rnn}& 	0.790&	0.667&	0.704&	0.685&		0.754	\\
		AttSiaBiLSTM~\cite{bao2018attentive}&0.695&	0.569&	0.554&	0.561&		0.658	\\
		LSTM+FNN+attention~\cite{chi2018sentence}& 0.818&	0.781&	0.715&	0.747&		0.809	\\
		CETE~\cite{laskar2020contextualized}& 0.874&	0.855&	0.795&	0.824&		0.870	\\
		STS-AM~\cite{moravvej2021method}& 0.756&	0.625&	0.650&	0.637&		0.718	\\ 
		\hline
		BPD+random weights& 0.808&	0.777&	0.698&	0.735&	0.801\\
		BPD+random weights+FL&	0.815&	0.784&	0.708&	0.744&	0.808\\
		
		BPD-IDE&	\textbf{0.930}&	\textbf{0.920}&	\textbf{0.881}&	\textbf{0.900}& \textbf{0.927}\\
		\hline
		\hline
	\end{tabular}
	\label{r1}
\end{table}

\begin{table}[ht]
	\centering
	\caption{Results of deep learning models on MSRP dataset.}
	\begin{tabular}{lccccc}
		\hline
		\hline
		& accuracy & recall & precision & F-measure & G-means \\
		\hline
		\hline
		RNN~\cite{sanborn2015deep}& 0.853&	0.922&	0.866&	0.893&		0.812	\\	
		Siamese CNN+LSTM~\cite{pontes2018predicting} & 0.863&	0.916&	0.882&	0.899&		0.833 	\\
		CA-RNN~\cite{chen2018rnn}& 0.880&	0.928&	0.896&	0.912&		0.854	\\
		AttSiaBiLSTM~\cite{bao2018attentive}& 0.874&	0.927&	0.889&	0.908&		0.845	\\
		LSTM+FNN+attention~\cite{chi2018sentence}& 0.889&	0.917&	0.916&	0.916&		0.873	\\
		CETE~\cite{laskar2020contextualized}& 0.916&	0.949&	0.926&	0.937&		0.898	\\
		STS-AM~\cite{moravvej2021method}& 0.899&	0.940&	0.910&	0.925&		0.876	\\
		\hline
		BPD+random weights& 0.875&	0.908&	0.905&	0.906&	0.858\\
		BPD+random weights+FL&	0.895&	0.926&	0.917&	0.921& 0.879\\
		BPD-IDE&	\textbf{0.937}&	\textbf{0.961}&	\textbf{0.946}&	\textbf{0.953}&	\textbf{0.925}\\
		\hline
		\hline
	\end{tabular}
	\label{r2}
\end{table}

\begin{table}[ht]
	\centering
	\caption{Results of deep learning models on SemEval2014 dataset.}
	\begin{tabular}{lccccc}
		\hline
		\hline
		& accuracy & recall & precision & F-measure & G-means \\
		\hline
		\hline
		RNN~\cite{sanborn2015deep}&	0.809&	0.822&	0.963&	0.887&	0.750\\
		Siamese CNN+LSTM~\cite{pontes2018predicting} & 0.775&	0.787&	0.958&	0.864&	0.720 	\\
		CA-RNN~\cite{chen2018rnn}& 0.811&	0.826&	0.961&	0.888&	0.742	\\
		AttSiaBiLSTM~\cite{bao2018attentive}& 0.799&	0.816&	0.957&	0.881&	0.720	\\
		LSTM+FNN+attention~\cite{chi2018sentence}& 0.733&	0.746&	0.949&	0.835&  0.670	\\
		CETE~\cite{laskar2020contextualized}& 0.854&	0.868&	0.969&	0.916&	0.791	\\
		STS-AM~\cite{moravvej2021method}&0.823&	0.834&	0.966&	0.895&	0.768	\\
		\hline
		BPD+random weights&	0.839&	0.855&	0.964&	0.906 &	0.766\\
		BPD+random weights+FL&	0.849&	0.863&	0.967&	0.912& 0.783\\
		BPD-IDE&	\textbf{0.876}&	\textbf{0.884}&	\textbf{0.977}&	\textbf{0.928}&\textbf{0.838}\\
		\hline
		\hline
	\end{tabular}
	\label{r3}
\end{table}


\subsubsection{Comparison with Other Metaheuristics}
In the next experiment, are our improved DE algorithm with a number of metaheuristic optimisation algorithms. That is, we use different metaheuristics to obtain the initial model parameters, while keeping the other model components, i.e., pre-processing, word embedding, LSTM and network structure, and loss function, the same. We use eight different algorithms, namely (standard) DE~\cite{price2013differential}, firefly algorithm (FA)~\cite{yang2010firefly}, bat algorithm (BA)~\cite{BAT_Main_Paper}, cuckoo optimisation algorithm (COA)~\cite{yang2009cuckoo},  artificial bee colony (ABC)~\cite{ABC_Main_Paper}, grey wolf optimisation (GWO)~\cite{mirjalili2014grey}, whale optimisation algorithm (WOA)~\cite{mirjalili2016whale}, and salp swarm algorithm (SSA)~\cite{bairathi2019salp} . For all algorithms, the population size and number of function evaluations are set to 200 and 3,000, respectively, while the default settings listed in Table~\ref{setting}. 

\begin{table}[h]
	\centering
	\caption{Metaheuristics parameter settings.}
	\label{table}
	\begin{tabular}{l|lc}
		\hline
		\hline
		algorithm & parameter  & value \\
		\hline
		\hline
		BPD-DE& scaling factor&	 0.5\\
		& crossover probability&	 0.8\\
		\hline
		BPD-FA& light absorption coefficient &	1\\
		& attractiveness at $r = 0$&	0.1\\	
		& scaling factor&	0.25\\
		\hline
		BPD-BA& constant for loudness update&	0.50\\
		& constant for an emission rate update&	0.50\\
		& initial pulse emission rate&	0.001\\
		\hline
		BPD-COA& discovery rate of alien solutions&	0.25\\
		\hline
		BPD-ABC & limit&	$n_e$ $\times$ dimensionality\\
		&$n_o$&	50\% of the colony\\		
		&$n_e$&	50\% of the colony\\
		&$n_s$&	1\\
		\hline
		BPD-GWO& no parameters\\
		\hline
		BPD-WOA& $b$&	1\\
		\hline
		BPD-SSA& no parameters\\
		\hline
		\hline
	\end{tabular}
	\label{setting}
\end{table}

For the SNLI dataset, BPD-IDE reduces error by about 31\% compared to the standard DE. It clearly shows that BPD-IDE has a substantial ability compared to the standard one. Also, DE offers more acceptable results than other algorithms, including ABC, GWO, and BAT. There is a minor improvement for the other two datasets, so the error rate for MSRP and SemEval2014 is reduced by around 16.07\% and 8.86\%, respectively.


\begin{table}[ht]
	\centering
	\caption{Results of metaheuristic algorithms on SNLI dataset.}
	\begin{tabular}{lccccc}
		\hline
		\hline
		& accuracy & recall & precision & F-measure & G-means \\
		\hline
		\hline
		BPD-DE& 0.897&	0.889&	0.824&	0.855&0.895\\
		BPD-FA& 0.864&	0.803&	0.801&	0.802& 0.848\\
		BPD-BA& 0.876&	0.850&	0.801&	0.825& 0.870\\
		BPD-COA& 0.860&	0.811&	0.787&	0.799 &0.847\\
		BPD-ABC & 0.885&	0.869&	0.809&	0.838& 0.881\\
		BPD-GWO& 0.842&	0.780&	0.763&	0.771&	0.826\\
		BPD-WOA& 0.883&	0.832&	0.828&	0.830& 0.870\\
		BPD-SSA& 0.863&	0.820&	0.789&	0.804& 0.852\\ \hline
		BPD-IDE & \textbf{0.930}&	\textbf{0.920}&	\textbf{0.881}&	\textbf{0.900}&		\textbf{0.927}	\\			
		\hline
		\hline
	\end{tabular}
	\label{r4}
\end{table}

\begin{table}[ht]
	\centering
	\caption{Results of metaheuristic algorithms on MSRP dataset.}
	\begin{tabular}{lccccc}
		\hline
		\hline
		& accuracy & recall & precision & F-measure & G-means \\
		\hline
		\hline
		BPD-DE& 0.925&	0.959&	0.930&	0.944& 0.906\\
		BPD-FA&0.897&	0.942&	0.908&	0.925&	0.874\\
		BPD-BA& 0.910&	0.944&	0.922&	0.933&	0.892\\
		BPD-COA& 0.902&	0.936&	0.918&	0.927& 0.884\\
		BPD-ABC & 0.899&0.946&	0.906&	0.926& 0.873\\
		BPD-GWO& 0.884&	0.926&	0.902&	0.914& 0.861\\
		BPD-WOA& 0.901&	0.938&	0.915&	0.926& 0.881\\
		BPD-SSA& 0.890&	0.929&	0.908&	0.918&	0.869\\ \hline
		BPD-IDE & \textbf{0.937}&	\textbf{0.961}&	\textbf{0.946}&	\textbf{0.953}&		\textbf{0.925}	\\			
		\hline
		\hline
	\end{tabular}
	\label{r5}
\end{table}

\begin{table}[ht]
	\centering
	\caption{Results of metaheuristic algorithms on SemEval2014 dataset.}
	\begin{tabular}{lccccc}
		\hline
		\hline
		& accuracy & recall & precision & F-measure & G-means \\
		\hline
		\hline
		BPD-DE& 0.864&	0.873&	0.975&	0.921&	0.822\\
		BPD-FA& 0.854&	0.865&	0.972&	0.915&   0.807\\
		BPD-BA&0.860&	0.869&	0.973&	0.918&	0.814 \\
		BPD-COA&0.856&	0.867&	0.971&	0.916&	0.805\\
		BPD-ABC & 0.851&	0.863&	0.970&	0.913&	0.796\\
		BPD-GWO&0.848&	0.857&	0.972&	0.911&0.805\\
		BPD-WOA&0.844&	0.856&	0.969&	0.909&	0.788\\
		BPD-SSA& 0.843&	0.857&	0.966&	0.908&	0.777\\\hline
		\textbf{BPD-IDE} & \textbf{0.876}&	\textbf{0.884}&	\textbf{0.997}&	\textbf{0.928}&		\textbf{0.838}	\\			
		\hline
		\hline
	\end{tabular}
	\label{r6}
\end{table}

\subsubsection{Word Embeddings}
We compare the effectiveness of the BERT model we employ in our approach for word embedding to five other word embeddings methods. One-Hot encoding~\cite{hackeling2017mastering} generates a binary feature for each class and assigns a value of one to each sample's features that correspond to the actual class. CBOW~\cite{sonkar2020attention} and Skip-gram~\cite{mccormick2016word2vec} use neural networks to map words to their embedding vectors. GloVe~\cite{pennington2014glove} is an unsupervised learning algorithm executed on a corpus's global word-word co-occurrence statistics. Instead of learning vectors for words, each word is represented as an n-gram of characters in FastText~\cite{thavareesan2020sentiment}, which is an extension of the Skip-gram model.

The results of this experiment are given in Tables~\ref{r70}, \ref{r80} and~\ref{r901} for the SNLI, MSRP, and SemEval2014 dataset, respectively. As expected, One Hot encoding has the worst performance among other word embeds, so in the MSRP dataset, as the best performance of this word embedding, the improvement of BPD-IDE for the two F-measure and G-means criteria are roughly 84.88\% and 82.51\%, respectively. CBOW and Skip-gram perform nearly identically in all three datasets due to their similar architecture, both of which are superior to the Glove word embedding. FastText acts best among other models but performs poorly on the BERT model. The BERT model decreases error by more than 11\%, 10\%, and 19\% compared to the FastText model for the SNLI, MSRP, and SemEval2014 datasets.


\begin{table}[ht]
	\centering
	\caption{Results of different word embeddings on SNLI dataset.}
	\begin{tabular}{lccccc}
		\hline
		\hline
		& accuracy & recall & precision & F-measure & G-means \\
		\hline
		\hline
		One-Hot encoding& 0.650&	0.473&	0.489&	0.481&	0.592\\	
		CBOW& 0.856&	0.779&	0.796&	0.787&	0.835\\	
		Skip-gram&0.871&	0.817&	0.808&	0.812&	0.857\\
		GloVe& 0.845&	0.798&	0.762&	0.780&	0.833\\ 
		FastText& 0.905&	0.861&	0.861&	0.861&	0.893\\
		\hline
		BERT & \textbf{0.930}&	\textbf{0.920}&	\textbf{0.881}&	\textbf{0.900}&		\textbf{0.927}	\\			
		\hline
		\hline
	\end{tabular}
	\label{r70}
\end{table}

\begin{table}[ht]
	\centering
	\caption{Results of different word embeddings on MSRP dataset.}
	\begin{tabular}{lccccc}
		\hline
		\hline
		& accuracy & recall & precision & F-measure & G-means \\
		\hline
		\hline	
		One-Hot encoding& 0.604&	0.659&	0.721&	0.689&	0.571\\
		CBOW& 0.802&	0.840&	0.859&	0.849&	0.781\\
		Skip-gram& 0.830&	0.856&	0.884&	0.870&	0.816\\	
		GloVe&0.781&	0.824&	0.844&	0.834& 0.758 \\
		FastText&0.864&	0.880&	0.913&	0.896&	0.857 \\
		\hline
		BERT & \textbf{0.937}&	\textbf{0.961}&	\textbf{0.946}&	\textbf{0.953}&		\textbf{0.925}	\\			
		\hline
		\hline
	\end{tabular}
	\label{r80}
\end{table}

\begin{table}[ht]
	\centering
	\caption{Results of different word embeddings on SemEval2014 dataset.}
	\begin{tabular}{lccccc}
		\hline
		\hline
		& accuracy & recall & precision & F-measure & G-means \\
		\hline
		\hline	
		One-Hot encoding& 0.504&	0.529&	0.875&	0.659&	0.367\\
		CBOW& 0.749&	0.768&	0.946&	0.848&	0.659\\
		Skip-gram& 0.758&	0.773&	0.951&	0.853&	0.686\\	
		Glove&0.697&	0.715&	0.936&	0.811&	0.606 \\
		FastText&0.812&	0.826&	0.961&	0.888&	0.742 \\
		\hline
		BERT & \textbf{0.876}&	\textbf{0.884}&	\textbf{0.997}&	\textbf{0.928}&		\textbf{0.838}	\\			
		\hline
	\end{tabular}
	\label{r901}
\end{table}

\begin{table}[ht]
	\centering
	\caption{Results of different loss functions on SNLI dataset.}
	\begin{tabular}{lccccc}
		\hline
		\hline
		& accuracy & recall & precision & F-measure & G-means \\
		\hline
		\hline
		WCE& 0.871&	0.856&	0.788&	0.821& 0.868\\
		BCE& 0.895&	0.874&	0.828&	0.850&	0.890\\
		DL& 0.915&	0.885&	0.870&	0.877& 0.908\\
		TL& 0.905&	0.880&	0.848&	0.864& 0.899\\ 
		\hline
		FL & \textbf{0.930}&	\textbf{0.920}&	\textbf{0.881}&	\textbf{0.900}&		\textbf{0.927}	\\			
		\hline
		\hline
	\end{tabular}
	\label{r7}
\end{table}

\begin{table}[ht]
	\centering
	\caption{Results of different loss functions on MSRP dataset.}
	\begin{tabular}{lccccc}
		\hline
		\hline
		& accuracy & recall & precision & F-measure & G-means \\
		\hline
		\hline
		WCE& 0.861&	0.906&	0.887&	0.896&	0.836\\
		BCE& 0.883&	0.923&	0.903&	0.913&	0.861\\
		DL& 0.915&	0.943&	0.930&	0.936&	0.899\\
		TL&0.899&	0.933&	0.917&	0.925&	0.881 \\ 
		\hline
		FL & \textbf{0.937}&	\textbf{0.961}&	\textbf{0.946}&	\textbf{0.953}&		\textbf{0.925}	\\			
		\hline
		\hline
	\end{tabular}
	\label{r8}
\end{table}

\begin{table}[ht]
	\centering
	\caption{Results of different loss functions on SemEval2014 dataset.}
	\begin{tabular}{lccccc}
		\hline
		\hline
		& accuracy & recall & precision & F-measure & G-means \\
		\hline
		\hline	
		WCE& 0.876 & \textbf{0.904} &	0.957 & \textbf{0.930} & 0.736\\
		BCE& 0.875&	0.899&	0.960&	0.928&	0.754\\
		DL& \textbf{0.877} &	0.890&	0.972&	0.929&	0.815\\
		TL& 0.876&	0.895&	0.966&	0.929&	0.784\\ 
		\hline
		FL & 0.876 & 0.884 & \textbf{0.997} & 0.928 & \textbf{0.838}	\\			
		\hline
		\hline
	\end{tabular}
	\label{r9}
\end{table}

\begin{table}[ht]
	\centering
	\caption{Top ranked suspicious sentences for source sentence "Two people are kickboxing and spectators are watching."  Words that appear in the source sentence are bolded.}
	\begin{tabular}{l|p{0.3\linewidth}|p{0.3\linewidth}|p{0.3\linewidth}}
		\hline
		\hline
		rank & BPD+random weights+FL &  BPD-IDE without FL &  BPD-IDE  \\
		\hline
		\hline
		1 & \textbf{Two people are} wading through the water & \textbf{Two people are} riding a motorcycle& \textbf{Two people are} fighting \textbf{and spectators are watching} \\
		\hline
		2 & A few men \textbf{are watching} cricket & \textbf{Two people are} fighting \textbf{and spectators are watching} & Two \textbf{spectators} \textbf{are kickboxing and} some people \textbf{are watching}\\
		\hline
		3 & Two girls are laughing breathlessly and other girls \textbf{are watching} them & Two adults are sitting in the chairs and \textbf{are watching} the ocean & Two young women are sparring in a \textbf{kickboxing} fight\\
		\hline
		4 & \textbf{Two people} wearing snowsuits are on the ground making snow angels &  Two \textbf{spectators} \textbf{are kickboxing and} some people \textbf{are watching} & Two women are \textbf{sparring} in a \textbf{kickboxing} match\\
		\hline	
		5 &   Two \textbf{spectators} \textbf{are kickboxing and} some people \textbf{are watching} &  A few men \textbf{are watching} cricket& \textbf{Two people are} wading through the water\\
		\hline
		\hline
	\end{tabular}
	\label{r90}
\end{table}

\subsubsection{Loss Functions}
Finally, to justify the select of focal loss in our BPD-IDE approach, we compare it to four other loss functions, weighted cross-entropy (WCE)~\cite{pihur2007weighted}, balanced cross-entropy (BCE)~\cite{xie2015holistically}, Dice loss (DL)~\cite{sudre2017generalised}, and Tversky loss (TL)~\cite{sadegh2017tversky}. 

The results of this experiments are given Tables~\ref{r7}, \ref{r8}, and~\ref{r9} for the SNLI, MSRP, and SemEval2014 dataset, respectively.  Generally speaking, the reduction of FL error compared to TL for SNLI and MSRP datasets is about 19\% and 27\%. However, these two functions are slightly different in the SemEval2014dataset, so the improvement rate for this dataset is about 12\%.


\subsubsection{Examples}
We give a qualitative example of the important contributions that both the improved DE and use of FL make to our approach based on the source sentence "Two people are kickboxing and spectators are watching" from the SemEval2014 dataset. Table~\ref{r90} gives the results of the top five sentences retrieved by the BPD model with random weight initialisation and focal loss, BPD-IDE without FL, and the full BPD-IDE approach. As is apparent, the full BPD-IDE model extracts suspicious sentences which are most similar to the source sentence, while the other two models retrieve these only in the lower rankings.

\section{Conclusions}
\label{Sec:conc}
In this paper, we have proposed a novel plagiarism detection model, founded on BERT word embedding, an attention mechanism-based LSTM method and an improved DE algorithm for pre-training the networks, while focal loss is used to address the inherent class imbalance. Our improved DE algorithm clusters the current population to find a promising region in search space and introduces a new updating strategy. Extensive experiments on three benchmark datasets confirm our approach to yield excellent performance, outperforming various other plagiarism detection approaches while the introduces DE algorithm is shown to be superior to several other metaheuristic algorithms.


Although the imbalanced classification problem can be partially solved by focal loss, additional approaches, such as reinforcement learning, are also available. In subsequent research, we will use reinforcement learning to address the imbalanced categorisation problem. As a workaround, we can think of each piece of data as a state and the research network as a policy. Additionally, we consider the minority class more than the majority class regarding rewards and penalties.
	
	In other words, the BERT model typically comprises millions of parameters, increasing the computing costs associated with these models' environmental scaling. This problem can be solved using other BERT extensions, like the DistilBERT language model. For the following work, we intend to use the other language models with fewer parameters.

\section*{Data Availability}
The data used to support the findings of this study are included within the article.
\section*{Conflicts of Interest}
Some of the authors are guest editors of the special issue.
\section*{Funding Statement}
This research received no specific grant from any funding agency in the public, commercial, or not-for-profit sectors.

\bibliography{References}
\bibliographystyle{plain}

\end{document}